\documentclass[lettersize,journal]{IEEEtran}
\usepackage{amsmath,amsfonts}
\usepackage{algorithmic}
\usepackage{algorithm}
\usepackage{array}
\usepackage[caption=false,font=normalsize,labelfont=sf,textfont=sf]{subfig}
\usepackage{textcomp}
\usepackage{stfloats}
\usepackage{url}
\usepackage{verbatim}
\usepackage{graphicx}
\usepackage{cite}
\usepackage{booktabs}
\usepackage{bm}
\usepackage{multirow}
\begin{document}
\title{DSGNN: A Dual-View Supergrid-Aware Graph Neural Network for Regional Air Quality Estimation}

\author{Xin Zhang, Ling Chen, Xing Tang, Hongyu Shi 
\thanks{Corresponding author: Ling Chen.}
\thanks{Xin Zhang, Ling Chen, Xing Tang, and Hongyu Shi are with the State Key Laboratory of Blockchain and Data Security, Zhejiang University, Hangzhou 310027, China, and also with the College of Computer Science and Technology, Zhejiang University, Hangzhou 310027, China (e-mail: xin.zhang@cs.zju.edu.cn; lingchen@cs.zju.edu.cn; tangxing@cs.zju.edu.cn; shihongyu@cs.zju.edu.cn).}
}

\markboth{Journal of \LaTeX\ Class Files,~Vol.~14, No.~8, March~2024}%
{Shell \MakeLowercase{\textit{et al.}}: A Sample Article Using IEEEtran.cls for IEEE Journals}


\maketitle

\nocite{wei2019satellite}
\nocite{arystanbekova2004application}
\nocite{kim2012urban, rakowska2014impact}
\nocite{huang2008cfd}
\nocite{hoek2008review, shad2009predicting}
\nocite{jutzeler2014region, patel2022accurate, li2014estimating}
\nocite{lin2017mining}
\nocite{lin2017mining, chen2022urban}
\nocite{patel2022accurate}
\nocite{chen2019deep}
\nocite{cheng2018neural}
\nocite{vaswani2017attention}
\nocite{han2021fine}
\nocite{chen2023deep}
\nocite{badrinarayanan2017segnet}
\nocite{song2022novel}
\nocite{dai2016r}

\begin{abstract}
Air quality estimation can provide air quality for target regions without air quality stations, which is useful for the public. Existing air quality estimation methods divide the study area into disjointed grid regions, and apply 2D convolution to model the spatial dependencies of adjacent grid regions based on the first law of geography, failing to model the spatial dependencies of distant grid regions. To this end, we propose a Dual-view Supergrid-aware Graph Neural Network (DSGNN) for regional air quality estimation, which can model the spatial dependencies of distant grid regions from dual views (i.e., satellite-derived aerosol optical depth (AOD) and meteorology). Specifically, images are utilized to represent the regional data (i.e., AOD data and meteorology data). The dual-view supergrid learning module is introduced to generate supergrids in a parameterized way. Based on the dual-view supergrids, the dual-view implicit correlation encoding module is introduced to learn the correlations between pairwise supergrids. In addition, the dual-view message passing network is introduced to implement the information interaction on the supergrid graphs and images. Extensive experiments on two real-world datasets demonstrate that DSGNN achieves the state-of-the-art performances on the air quality estimation task, outperforming the best baseline by an average of 19.64\% in MAE.
\end{abstract}

\begin{IEEEkeywords}
Deep learning, graph neural network, multi-task
learning, air quality estimation.
\end{IEEEkeywords}

\section{Introduction}

With the rapid development of industrialization and urbanization, air pollution has become an increasingly serious problem. The high levels of air pollution concentrations pose a significant threat to human health. The air quality data of regions can be accurately obtained by air quality stations. Due to the high cost of building and maintaining these stations, the number of air quality stations is limited, which are primarily concentrated in urban areas. As a result, air quality estimation for target regions without air quality stations is useful for the public.

Existing air quality estimation methods can be classified into two categories: dispersion model based methods and data-driven methods. Dispersion model based methods try to estimate air quality by simulating the diffusion of air pollutants. These methods employ dispersion models (e.g., Gaussian plume model \cite{arystanbekova2004application}, street canyon model \cite{kim2012urban, rakowska2014impact}, computational fluid dynamics model \cite{huang2008cfd}, and transport-chemical model \cite{carnevale2009neuro}) to simulate the diffusion of air pollutants, which are based on the theories of air motion and matter diffusion. However, the parameters used in these dispersion models are set by experience, and the empirical assumptions cannot accurately reflect the real situation.

Data-driven methods try to estimate air quality by modeling the relevance between air quality and input data. Early data-driven methods employ statistical regression models (e.g., linear regression model \cite{hoek2008review, shad2009predicting}, Gaussian process regression model \cite{jutzeler2014region, patel2022accurate}, and generalized additive model \cite{lin2017mining}) to capture the relevance with input data, mainly including meteorology data, road network data, land cover data, etc. With the advancements in remote sensing, many institutions have released satellite-derived aerosol optical depth (AOD) data, which are the integration of aerosol extinction in the total atmospheric column and have a strong correlation with air pollution concentrations \cite{lv2017daily,jung2021national,goldberg2019using}. Consequently, researchers incorporate AOD data into statistical regression models to estimate air quality \cite{acp2032732020, geng2021trackingair,jiang2021estimation}. Recently, encouraged by the success of deep neural networks (DNN), some researchers employ DNN to automate feature learning \cite{cheng2018neural, chen2019deep, han2021fine}. These methods employ graph embedding \cite{chen2019deep}, graph neural networks (GNN) \cite{han2021fine, han2022semi, han2023kill}, and self-attention networks \cite{cheng2018neural, han2021fine} to model the spatial dependencies between the target region and other regions with air quality stations. However, these methods estimate the air quality of target regions separately, which fail to model the spatial dependencies among target regions.


To address the problem, some researchers divide the study area into disjointed grid regions, use convolutional neural networks (CNN) to model the complex spatial dependencies of grid regions (i.e., target regions and regions with air quality stations), and estimate the air quality of all target regions at a time \cite{chen2023deep, song2022novel}. These methods apply 2D convolution to model the spatial dependencies of adjacent grid regions based on the first law of geography. In fact, some distant grid regions also have strong spatial dependencies, e.g., two distant coastal grid regions that are both affected by sea breezes.

To address the above-mentioned drawback, we propose a Dual-view Supergrid-aware Graph Neural Network (DSGNN) for regional air quality estimation. DSGNN groups correlated grid regions into supergrids from dual views, and constructs the supergrid graphs to model the spatial dependencies of supergrids, which can model the spatial dependencies of both adjacent and distant grid regions.

The main contributions of this paper are as follows:
\begin{itemize}
    \item We introduce a dual-view supergrid learning (DSL) module to generate supergrids from dual views (i.e., AOD and meteorology) in a parameterized way, which can group correlated grid regions into supergrids.
    \item We introduce a dual-view implicit correlation encoding (DCE) module, which can learn the correlations between pairwise supergrids.
    \item We introduce a dual-view message passing (DMP) network to implement the information interaction on the dual-view supergrid graphs and images, which can model the spatial dependencies of both adjacent and distant grid regions.
    \item We collect two datasets (i.e., YRD-AOD and BTH-AOD), including not only the commonly used data (i.e., meteorology data and air quality data) but also AOD data. The experimental results on the two datasets demonstrate that DSGNN achieves the state-of-the-art (SOTA) performance in regional air quality estimation, outperforming the best baseline by an average of 19.64\% in MAE.
\end{itemize}

\section{Related Work}

This section provides an overview of the related work, including dispersion model based methods and data-driven methods.
\subsection{Dispersion Model Based Methods}
Dispersion model based methods try to estimate air quality by simulating the diffusion of air pollutants in the atmosphere, which rely on mathematical theories that describe the physical and chemical processes governing the movement and transformation of air pollutants. For example, Arystanbekova et al. \cite{arystanbekova2004application} employed the Gaussian plume model to simulate the physical diffusion and chemical reactions of air pollutants. Kim et al. \cite{kim2012urban} and Rakowska et al. \cite{rakowska2014impact} utilized the street canyon model to capture the impact of vehicle emitted pollutants on air quality by considering traffic conditions. Huang et al. \cite{huang2008cfd} used the three-dimensional computational fluid dynamics model to estimate air quality by simulating the variation of pollutant concentrations with wind speed and time. Li et al. \cite{li2022effects} introduced a computational fluid dynamic simulation based model with consideration of solar radiation, which captures the effects of both horizontal and vertical building setbacks on air quality. However, these dispersion models rely on empirical assumptions and expert knowledge to set the parameters, and the empirical assumptions cannot accurately reflect the real situation.

\subsection{Data-Driven Methods}

Data-driven methods try to estimate air quality by modeling the relevance between air quality and input data, which can be mainly divided into two categories: statistical regression model based methods and deep learning based methods.

Early statistical regression model based methods employ linear models (e.g., fuzzy genetic linear membership kriging model \cite{shad2009predicting, shukla2020mapping} and linear regression model \cite{hoek2008review, shad2009predicting}) to capture the relevance, but these models cannot exploit the nonlinear relevance. To address the problem, some researchers use Gaussian process regression model \cite{jutzeler2014region, patel2022accurate, li2014estimating}, generalized additive model \cite{lin2017mining}, and Random Forest model \cite{lin2017mining, chen2022urban} to capture the nonlinear relevance. For example, Patel et al. \cite{patel2022accurate} proposed Gaussian process based models to implicitly provide uncertainty and incorporate domain-specific information for air quality estimation. However, these methods are prone to overfitting due to the limited amount of labeled data. This problem leads to semi-supervised learning based methods \cite{zheng2013u, chen2016spatially, lv2019air} that can utilize the unlabeled data to assist training. For example, Zheng et al. \cite{zheng2013u} exploited co-training \cite{blum1998combining} combined with semi-supervised learning to estimate regional air quality. Chen et al. \cite{chen2016spatially} proposed an ensemble semi-supervised learning and pruning model to estimate urban air quality. Lv et al. \cite{lv2019air} proposed a transfer learning \cite{pan2009survey} based semi-supervised learning model to estimate air quality. However, the performance of these models depends on feature engineering, which is limited by human knowledge and may cause the loss of information.

Inspired by the success of DNN, some researchers employ DNN to automate feature learning. These methods employ graph embedding, GNN, and self-attention networks to model the spatial dependencies between the target region and other regions with air quality stations. For example, Chen et al. \cite{chen2019deep} proposed PANDA, which employs graph embedding \cite{cai2018comprehensive} to model the spatial dependencies between the target region and other regions with air quality stations, and jointly implements air quality estimation and air quality forecasting tasks. Cheng et al. \cite{cheng2018neural} proposed ADAIN, which employs self-attention networks \cite{vaswani2017attention} to learn the weights of different regions with air quality stations, and utilizes feedforward neural networks (FNN) to automatically learn the features. Han et al. \cite{han2021fine} proposed MCAM, which introduces a multi-channel GNN \cite{kipf2016semi} to model the dynamic and static spatial dependencies between the target region and other regions with air quality stations. Bonet et al. \cite{bonet2022explaining} proposed TNSGNN, which introduces a topology-aware kernelized node selection strategy to select the most relevant regions with air quality stations for each target region, and employs GNN to model the influence of the most relevant regions. Han et al. \cite{han2023kill} proposed $\text{MasterGNN}^+$, which employs a multi-view GNN to model the spatial dependencies between the target region and other regions with air quality stations from the geographical distance view and the environmental context view. However, these methods estimate the air quality of target regions separately, which fail to model the spatial dependencies among target regions.

To address the problem, some researchers divide the study area into disjointed grid regions, use convolutional neural networks (CNN) to model complex spatial dependencies of grid regions (i.e., target regions and regions with air quality stations), and estimate the air quality of all target regions at a time. For example, Zhang et al. \cite{zhang2022deep} proposed Deep-AIR, which constructs images from meteorology data, factory air pollutant emission data, and road network data, and introduces AirRes module to model the spatial dependencies of grid regions and provide multi-source feature interactions. Song et al. \cite{song2022novel} proposed Multi-AP, which constructs images from meteorology data, traffic data, and land cover data, and utilizes fully convolutional networks (FCN) \cite{dai2016r} to model the spatial dependencies of grid regions from micro, meso, and macro views. Chen et al. \cite{chen2023deep} proposed FAIRY, which constructs images from meteorology data, traffic data, factory air pollutant emission data, and point of interest data, employs SegNet \cite{badrinarayanan2017segnet} to model the spatial dependencies of grid regions, and uses self-attention networks to provide multi-source feature interactions. However, these methods apply 2D convolution to model the spatial dependencies of adjacent grid regions based on the first law of geography, which fail to model the spatial dependencies of distant grid regions.

In our work, the dual-view supergrid modeling is introduced to address the above-mentioned drawback. Empowered by this, DSGNN can model the spatial dependencies of both adjacent and distant grid regions.

\section{Preliminaries}
In this section, we provide the definitions of the associated terms used in DSGNN, and formulate the air quality estimation task.
\begin{figure}[h]
\centering
\includegraphics[width=0.9\columnwidth]{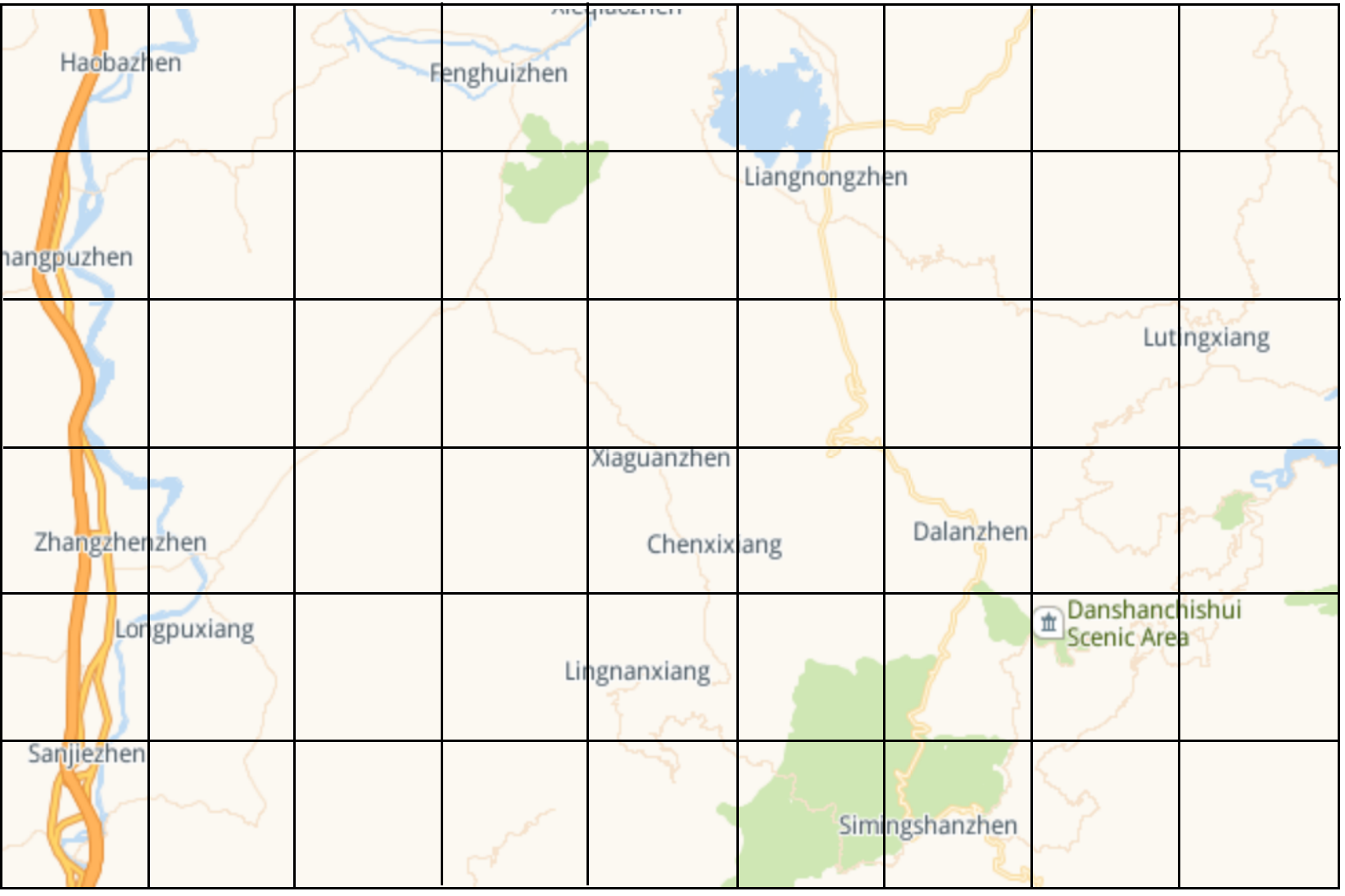} 
\caption{Example of grid regions. (Best viewed in color).}
\label{fig1}
\end{figure}

\textbf{Definition 1. Grid region.} As shown in Figure \ref{fig1}, the study area is divided into disjointed grid regions, assuming that the air quality in a grid region is uniform. The grid regions are denoted as $G$. $g_{i j}$ represents the grid region of the $i$-th row and the $j$-th column. The numbers of grid regions in the vertical and horizontal directions of the study area are $H$ and $W$, respectively.

\textbf{Definition 2. AOD image.} AOD image $\boldsymbol{I}_{\text{AOD}}^t$ contains one channel, which is constructed based on AOD data (i.e., the integration of aerosol extinction in the total atmospheric column). The AOD sequence at time step $t$ is denoted as $\boldsymbol{H}^t_{\text{AOD}}=\{{\boldsymbol{I}_{\text{AOD}}^{t-\tau+1}},{\boldsymbol{I}_{\text{AOD}}^{t-\tau+2}},\dots,{\boldsymbol{I}_{\text{AOD}}^{t}}\}$, where $\tau$ is the historical window length. The total AOD sequence is denoted as $\boldsymbol{\mathcal{H}}_{\text{AOD}}=\{{\boldsymbol{I}_{\text{AOD}}^{1}},{\boldsymbol{I}_{\text{AOD}}^{2}},\dots,{\boldsymbol{I}_{\text{AOD}}^{T}}\}$, where $T$ is the number of time steps observed during the training phase.

\textbf{Definition 3. Meteorology image.} Meteorology image $\boldsymbol{I}_{\text{Meteorology}}^t$ contains five channels, which is constructed based on meteorology data (i.e., temperature, humidity, rainfall, wind force, and wind direction). Similar to Definition 2, the meteorology sequence at time step $t$ and the total meteorology sequence are denoted as $\boldsymbol{H}^t_{\text{Meteorology}}$ and $\boldsymbol{\mathcal{H}}_{\text{Meteorology}}$, respectively.


\textbf{Definition 4. Air quality image.} Air quality image $\boldsymbol{I}_{\text{Air quality}}^t$ contains six channels, which is constructed based on air quality data (i.e., $\text{SO}_{2}$, $\text{NO}_{2}$, $\text{PM}_{10}$, $\text{CO}$, $\text{O}_{3}$, and $\text{PM}_{2.5}$). Similar to Definition 2, the air quality sequence at time step $t$ is denoted as $\boldsymbol{H}^t_{\text{Air quality}}$. Since some grid regions do not have air quality stations, the air quality data of these grid regions are filled in by a linear interpolation method\cite{chen2023deep}. If a grid region has more than one air quality station, the average air quality data of these air quality stations are used as the values of the grid region\cite{chen2023deep, song2022novel}.


\textbf{Problem 1. Air quality estimation task.} Given $\boldsymbol{H}^t_{\text{AOD}}$, $\boldsymbol{H}^t_{\text{Meteorology}}$, $\boldsymbol{H}^t_{\text{Air quality}}$, $\boldsymbol{\mathcal{H}}_{\text{AOD}}$, and $\boldsymbol{\mathcal{H}}_{\text{Meteorology}}$, the air quality estimation task aims to get ${\widetilde{\boldsymbol{I}}^t_{\text{Air quality}}}$, a single channel image containing one kind of air quality data (i.e., $\text{SO}_{2}$, $\text{NO}_{2}$, $\text{PM}_{10}$, $\text{CO}$, $\text{O}_{3}$, or $\text{PM}_{2.5}$) at time step $t$.



\section{Methodology}
\begin{figure*}[h]
    \centering
    \includegraphics[width=1\textwidth]{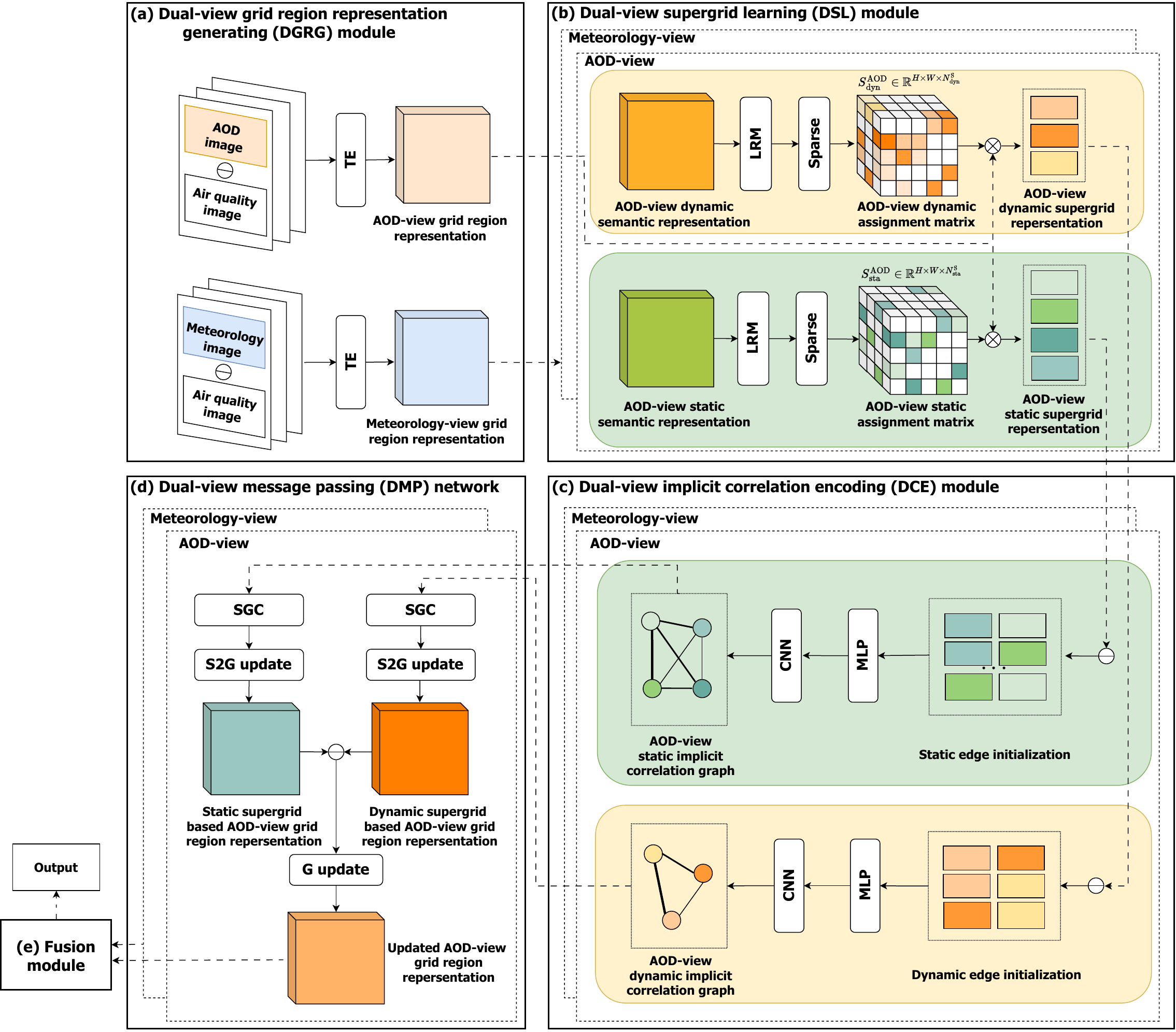} 
    \caption{Framework of DSGNN. \textit{TE} denotes the temporal encoder. \textit{LRM} denotes the low-rank-based mapper. \textit{SGC} denotes supergrid graph convolution. \textit{S2G update} denotes grid region representation updating based on supergrid representation. \textit{G update} denotes grid region representation updating. $\ominus$ denotes concatenation. $\otimes$ denotes multiplying. (Best viewed in color).}
    \label{fig2}
\end{figure*}

The framework of DSGNN is shown in Figure \ref{fig2}, which consists of five modules: (1) the dual-view grid region representation generating module employs the temporal encoder to generate the dual-view initial grid region representations; (2) the dual-view supergrid learning module utilizes assignment matrices to capture the mapping between grid regions and supergrids in a parameterized way, which groups correlated grid regions into supergrids; (3) the dual-view implicit correlation encoding module captures implicit correlations between pairwise supergrids; (4) the dual-view message passing network accomplishes the message aggregation and representation updating on the supergrid graphs and images, which can model the spatial dependencies of both adjacent and distant grid regions; (5) the fusion module fuses the dual-view grid region representations to estimate the air quality of all grid regions at a time.

\begin{figure}[h]
\centering
\includegraphics[width=1\columnwidth]{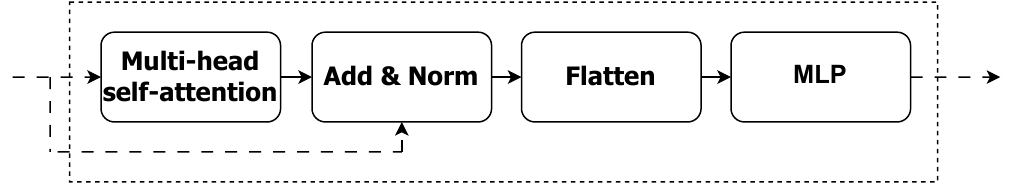} 
\caption{Architecture of the temporal encoder.}
\label{TE结构图}
\end{figure}

\subsection{Dual-View Grid Region Representation Generating Module}
To model the different spatial dependences of grid regions, the initial grid region representations are obtained from dual views (i.e., AOD and meteorology).


The temporal encoder is customized to obtain the initial grid region representations. The overall architecture of the temporal encoder is shown in Figure \ref{TE结构图}. After the point-wise attention operation implemented by matrix multiplication, a sequence of vectors is obtained. After that, the sequence of vectors is flattened to a 1-D vector, and then fed into the multi-layer perceptron (MLP) to compact the fine-grained information across time steps. For the AOD-view, the AOD sequence and air quality sequence are concatenated as input for each grid region. Then, the temporal encoder is utilized to obtain the initial AOD-view grid region representation $\boldsymbol{X}^\text{AOD}$, formulated as:
\begin{align}
\begin{split}
\boldsymbol{X}^\text{AOD}_{ij} = \text{TE}(\text{Concat}((\boldsymbol{H}^t_{\text{AOD}})_{ij},(\boldsymbol{H}^t_{\text{Air quality}})_{ij})) \label{eq:fun_f}
\end{split}
\end{align}
where $(\boldsymbol{H}^t_{\text{AOD}})_{ij}$ and $(\boldsymbol{H}^t_{\text{Air quality}})_{ij}$ denote the AOD sequence and air quality sequence of grid region $g_{i j}$, respectively, $i \in \{1,2,…,H\}$ and $j \in \{1,2,…,W\}$. $\text{Concat}(\cdot)$ denotes the concatenation operation. $\text{TE}(\cdot)$ denotes the temporal encoder. In the same way, the initial meteorology-view grid region representation $\boldsymbol{X}^{\text{Meteorology}}$ can be obtained.

\subsection{Dual-View Supergrid Learning Module}

Distant grid regions may have strong spatial dependencies. The spatial dependencies inherently have both dynamic and static aspects \cite{han2021fine,chen2022aargnn}. For the dynamic aspect, the spatial dependencies between grid regions are dependent on some dynamic properties and change over time. For example, with the change of weather conditions, the spatial dependencies between grid regions are different over time. For the static aspect, the spatial dependencies between grid regions are dependent on some static properties and stay unchanged. For example, due to the locational property, coastal grid regions all have a monsoon climate and exhibit strong spatial dependencies. Inspired by this, the dual-view supergrid learning module is designed to group correlated grid regions into dynamic and static supergrids in a learning way, which utilizes assignment matrices to capture the mapping between grid regions and supergrids. In real scenarios, each grid region may belong to multiple supergrids owing to different properties. 
Thus, instead of assigning each grid region into a specific supergrid, the dual-view supergrid learning module makes each grid region belong to multiple supergrids with different probabilities and the total probability equals to one.

Directly learning dense assignment matrices will introduce too many parameters and make the model hard to train, especially when the number of grid regions is large. To solve this problem, the low-rank-based mapper is introduced to obtain the assignment matrix in a learning way, which decomposes the assignment matrix into two low-rank matrices \cite{10184800, yang2022multi}. For the AOD-view, before obtaining assignment matrices, the temporal encoder is utilized to calculate the AOD-view dynamic semantic representation $\boldsymbol{E}_\text{dyn}^{\text{AOD}} \in R^{H \times W \times d} $ based on $\boldsymbol{H}^t_{\text{AOD}}$, where $d$ is the dimension of the representation, formulated as:
\begin{align}
    \begin{split}
    & (\boldsymbol{E}_\text{dyn}^\text{AOD})_{ij} = \text{TE}((\boldsymbol{H}^t_{\text{AOD}})_{ij})
    \label{eq:fun_f}
\end{split}
\end{align}
where $\text{TE}(\cdot)$ denotes the temporal encoder. The parameters of this temporal encoder are not shared with the temporal encoder in the AOD-view grid region representation generating module. Then, matrix factorization \cite{yu2016temporal, sen2019think} is utilized to calculate the AOD-view static semantic representation $\boldsymbol{E}_\text{sta}^{\text{AOD}} \in R^{H \times W \times d} $ based on the total AOD sequence $\boldsymbol{\mathcal{H}}_{\text{AOD}}$, as matrix factorization is capable of preserving as much information from the original sequences as possible when the sequence length is large \cite{6058636, yu2016temporal}. Subsequently, the AOD-view assignment matrices $\boldsymbol{S}^\text{AOD}_\text{dyn} \in R^{H \times W \times N_\text{dyn}^{\text{S}}}$ and $\boldsymbol{S}^\text{AOD}_\text{sta} \in R^{H \times W \times N_\text{sta}^{\text{S}}}$ are obtained, where $N_\text{dyn}^{\text{S}}$ and $N_\text{sta}^{\text{S}}$ are the numbers of dynamic and static supergrids, respectively, formulated as:
\begin{align}
\begin{split}
& \boldsymbol{S}^\text{AOD}_\text{dyn} =  \boldsymbol{E}_\text{dyn}^\text{AOD}\boldsymbol{W}_\text{dyn}^{\text{AOD}}\\
& \boldsymbol{S}^\text{AOD}_\text{sta} =  \boldsymbol{E}_\text{sta}^\text{AOD}\boldsymbol{W}_\text{sta}^{\text{AOD}}
\label{eq:fun_f}
\end{split}
\end{align}
where $\boldsymbol{W}_\text{dyn}^{\text{AOD}} \in R^{d \times N_\text{dyn}^{\text{S}}}$ and $\boldsymbol{W}_\text{sta}^{\text{AOD}} \in R^{d \times N_\text{sta}^{\text{S}}}$ are learnable parameters. 

To make the model more robust and reduce the impact of noise, we employ a sparse threshold strategy \cite{zhang2021hierarchical} to make $\boldsymbol{S}^\text{AOD}_\text{dyn}$ and $\boldsymbol{S}^\text{AOD}_\text{sta}$ sparse, which preserves the values above the threshold $\rho$, and the other values are truncated to 0. Similarly, the meteorology-view assignment matrices $\boldsymbol{S}^\text{Meteorology}_\text{dyn} \in R^{H \times W \times N_\text{dyn}^{\text{S}}}$ and  $\boldsymbol{S}^\text{Meteorology}_\text{sta} \in R^{H \times W \times N_\text{sta}^{\text{S}}}$ can be obtained.

Afterward, the AOD-view supergrid representations $\boldsymbol{Z}_\text{dyn}^\text{AOD}$ and $\boldsymbol{Z}_\text{sta}^\text{AOD}$ are obtained based on the initial grid region representation and assignment matrices, formulated as:
\begin{align}
\begin{split}
(\boldsymbol{Z}_\text{dyn}^\text{AOD})_{k} &=  \sum_{i=1}^H \sum_{j=1}^W {(\boldsymbol{S}^\text{AOD}_\text{dyn})_{ij{k}}} \boldsymbol{X}^\text{AOD}_{ij} \\
(\boldsymbol{Z}_\text{sta}^\text{AOD})_{l} &=  \sum_{i=1}^H \sum_{j=1}^W {(\boldsymbol{S}^\text{AOD}_\text{sta})_{ij{l}}} \boldsymbol{X}^\text{AOD}_{ij}
\label{eq:fun_f}
\end{split}
\end{align}
where $(\boldsymbol{S}^\text{AOD}_\text{dyn})_{ij{k}}$ is the probability of assigning grid region $g_{ij}$ to the $k$-th dynamic supergrid, $i \in \{1,2,…,H\}$, $j \in \{1,2,…,W\}$, and $k \in \{1,2,…,N_\text{dyn}^\text{S}\}$. $(\boldsymbol{S}^\text{AOD}_\text{sta})_{ij{l}}$ is the probability of assigning grid region $g_{ij}$ to the $l$-th static supergrid, $l \in \{1,2,…,N_\text{sta}^\text{S}\}$. In the same way, the meteorology-view supergrid representations $\boldsymbol{Z}_\text{dyn}^\text{Meteorology}$ and $\boldsymbol{Z}_\text{sta}^\text{Meteorology}$ can be obtained.

\subsection{Dual-View Implicit Correlation Encoding Module}

The dual-view implicit correlation encoding module is introduced to capture the implicit correlations between pairwise supergrids. To avoid missing any relevant information between pairwise supergrids, the dynamic and static supergrid graphs are modeled as fully connected graphs. For the AOD-view, the representations of the implicit correlations between pairwise supergrids are obtained based on the supergrid representations, formulated as:
\begin{align}
\begin{split}
(\boldsymbol{c}_\text{dyn}^\text{AOD})_{{i}{j}} &= \text{ReLU}\left(\phi_1(\text{Concat}((\boldsymbol{Z}_\text{dyn}^\text{AOD})_{i}, (\boldsymbol{Z}_\text{dyn}^\text{AOD})_{j}))\right) \\
(\boldsymbol{c}_\text{sta}^\text{AOD})_{{k}{l}} &= \text{ReLU}(\phi_2(\text{Concat}((\boldsymbol{Z}_\text{sta}^\text{AOD})_{k}, (\boldsymbol{Z}_\text{sta}^\text{AOD})_{l})))
\label{eq:fun_f}
\end{split}
\end{align}
where $(\boldsymbol{Z}_\text{dyn}^\text{AOD})_{i}$ and $(\boldsymbol{Z}_\text{dyn}^\text{AOD})_{j}$ are the AOD-view dynamic supergrid representations, and ${i},{j} \in \{1,2,…,N_\text{dyn}^\text{S}\}$. $(\boldsymbol{Z}_\text{sta}^\text{AOD})_{k}$ and $(\boldsymbol{Z}_\text{sta}^\text{AOD})_{l}$ are the AOD-view static supergrid representations, and ${k},{l} \in \{1,2,…,N_\text{sta}^\text{S}\}$. $\phi_1$ and $\phi_2$ are transformation functions, implemented by the MLP. Then, the weights of the implicit correlations can be obtained, formulated as:
\begin{align}
\begin{split}
({q}_\text{dyn}^\text{AOD})_{{i}{j}} &= \text{Sigmoid}(\text{Conv}_1((\boldsymbol{c}_\text{dyn}^\text{AOD})_{{i}{j}})) \\
({q}_\text{sta}^\text{AOD})_{{k}{l}} &= \text{Sigmoid}(\text{Conv}_2((\boldsymbol{c}_\text{sta}^\text{AOD})_{{k}{l}}))
\label{eq:fun_f}
\end{split}
\end{align}
where $\text{Conv}_1(\cdot)$ and $\text{Conv}_2(\cdot)$ denote 1D convolution layers. $\text{Sigmoid}(\cdot)$ is the Sigmoid activation function, which keeps the value of the weight ranging from 0 to 1. In the same way, the meteorology-view weights of the implicit correlations ${q}_\text{dyn}^\text{Meteorology}$ and ${q}_\text{sta}^\text{Meteorology}$ can be obtained.

\subsection{Dual-View Message Passing Network}

Supergrid graph convolution is customized to model the interactions of supergrids, which includes the aggregating operation and the updating operation. In the aggregating operation, the stronger the weight of the implicit correlation between a supergrid pair is, the more critical the implicit correlation is in the aggregating operation and vice versa. Thus, we aggregate messages from connected edges based on the weights of the implicit correlations. For the AOD-view, the aggregating operation can be formulated as:
\begin{align}
\begin{split}
(\boldsymbol{r}_\text{dyn}^\text{AOD})_i =  \sum_{i\neq j}({q}_\text{dyn}^\text{AOD})_{ij} (\boldsymbol{c}_\text{dyn}^\text{AOD})_{ij} \\
(\boldsymbol{r}_\text{sta}^\text{AOD})_k =  \sum_{k\neq l}({q}_\text{sta}^\text{AOD})_{kl} (\boldsymbol{c}_\text{sta}^\text{AOD})_{kl}
\label{eq:fun_f}
\end{split}
\end{align}
where $({q}_\text{dyn}^\text{AOD})_{ij}$ and $({q}_\text{sta}^\text{AOD})_{kl}$ are the weights of the implicit correlations calculated by Equation (6). $(\boldsymbol{c}_\text{dyn}^\text{AOD})_{ij}$ and $(\boldsymbol{c}_\text{sta}^\text{AOD})_{kl}$ are the representations of the implicit correlations calculated by Equation (5). In the updating operation, we obtain the updated AOD-view supergrid representations ${\boldsymbol{Z}_\text{dyn}^\text{AOD}}'$ and ${\boldsymbol{Z}_\text{sta}^\text{AOD}}'$, formulated as:
\begin{align}
\begin{split}
{{{(\boldsymbol{Z}_\text{dyn}^\text{AOD}}')_i}} &=  \phi_3((\boldsymbol{Z}_\text{dyn}^\text{AOD})_i,(\boldsymbol{r}_\text{dyn}^\text{AOD})_i) \\
{{{({\boldsymbol{Z}_\text{sta}^\text{AOD}}')}_k}} &=  \phi_4((\boldsymbol{Z}_\text{sta}^\text{AOD})_k,(\boldsymbol{r}_\text{sta}^\text{AOD})_k)
\label{eq:fun_f}
\end{split}
\end{align}
where $\phi_3$ and $\phi_4$ are transformation functions, implemented by the MLP. Similarly, the updated meteorology-view supergrid representations ${\boldsymbol{Z}_\text{dyn}^\text{Meteorology}}'$ and ${\boldsymbol{Z}_\text{sta}^\text{Meteorology}}'$ can be obtained.

To implement the information interaction from the supergrid graphs to images, \textit{S2G update} (grid region representation updating based on supergrid representation) is customized to obtain the supergrid based grid region representations. For the AOD-view, \textit{S2G update} utilizes assignment matrices to obtain the supergrid based AOD-view grid region representations ${\boldsymbol{X}_\text{dyn}^\text{AOD}}'$ and ${\boldsymbol{X}_\text{sta}^\text{AOD}}'$, formulated as:
\begin{align}
\begin{split}
{{({\boldsymbol{X}_\text{dyn}^\text{AOD}}')_{ij}}} &=  \sum_{k=1}^{N_\text{dyn}^\text{S}} {(\boldsymbol{S}_\text{dyn}^\text{AOD})_{ijk}} {{({\boldsymbol{Z}_\text{dyn}^\text{AOD}}')_{k}}} \\
{{({\boldsymbol{X}_\text{sta}^\text{AOD}}')_{ij}}} &=  \sum_{l=1}^{N_\text{sta}^\text{S}} {(\boldsymbol{S}_\text{sta}^\text{AOD})_{ijl}} {{({\boldsymbol{Z}_\text{sta}^\text{AOD}}')_{l}}}
\label{eq:fun_f}
\end{split}
\end{align}
where $N_\text{dyn}^\text{S}$ and $N_\text{sta}^\text{S}$ are the numbers of the dynamic and static supergrids, respectively. ${\boldsymbol{Z}_\text{dyn}^\text{AOD}}'$ and ${\boldsymbol{Z}_\text{sta}^\text{AOD}}'$ are the updated AOD-view supergrid representations, calculated by Equation (8). Similarly, the supergrid based meteorology-view grid region representations ${\boldsymbol{X}_\text{dyn}^\text{Meteorology}}'$ and ${\boldsymbol{X}_\text{sta}^\text{Meteorology}}'$ can be obtained.

\textit{G update} (grid region representation updating) is designed to model the spatial dependencies of both adjacent and distant grid regions. For the AOD-view, the updated AOD-view grid region representation ${\boldsymbol{X}^\text{AOD}}''$ is calculateed based on ${\boldsymbol{X}_\text{dyn}^\text{AOD}}'$, ${\boldsymbol{X}_\text{sta}^\text{AOD}}'$, and $\boldsymbol{X}^\text{AOD}$, formulated as:
\begin{align}
    {\boldsymbol{X}^\text{AOD}}'' =  \text{F}_\text{G update}(\text{Concat}({\boldsymbol{X}_\text{dyn}^\text{AOD}}', {\boldsymbol{X}_\text{sta}^\text{AOD}}',\boldsymbol{X}^\text{AOD}))
    \label{eq:fun_f}
\end{align}
where $\text{F}_\text{G update} (\cdot)$ is built with several 3×3 convolution layers followed by a batch normalization layer and a ReLU activation function. Similarly, the updated meteorology-view grid region representation ${\boldsymbol{X}^\text{Meteorology}}''$ can be obtained.

\subsection{Fusion Module}

The updated grid region representations are integrated to obtain the fused dual-view grid region representation. Then, we estimate the air quality of all grid regions at a time, formulated as:
\begin{align}
{\boldsymbol{X}^\text{F}} =  \alpha{\boldsymbol{X}^\text{AOD}}''+(1 - \alpha){\boldsymbol{X}^\text{Meteorology}}''
\label{eq:fun_f}
\end{align}
\begin{align}
{\widetilde{\boldsymbol{I}}^t_{\text{Air quality}}} =  \text{F}_\text{est}({\boldsymbol{X}^\text{F}})
\label{eq:fun_f}
\end{align}
where $\text{F}_\text{est} (\cdot)$ is an estimation function implemented by two $1 \times 1$ convolution layers. $\alpha$ is the importance weight of the updated AOD-view grid region representation, and $0\leq\alpha\leq1$.

\subsection{Learning}
To ensure that the grid regions with similar semantic representations are grouped into the same supergrid, the dual-view reconstruction loss $L_\text{recon}$ is introduced as a constraint in supergrid learning. Specifically, assignment matrices are utilized to reconstruct the semantic representations, and the distances between the reconstructed and corresponding input semantic representations should be as small as possible. For the AOD-view, the AOD-view reconstruction loss $L^\text{AOD}_\text{recon}$ is calculated based on the AOD-view assignment matrices and semantic representations, formulated as:
\begin{align}
\begin{split}
(L^\text{AOD}_\text{recon})_\text{dyn} & = \text{dist}\left(\boldsymbol{S}^\text{AOD}_\text{dyn}(\boldsymbol{S}^\text{AOD}_\text{dyn})^T\boldsymbol{E}_\text{dyn}^\text{AOD}, \boldsymbol{E}_\text{dyn}^\text{AOD}\right) \\
\vspace{+30pt}
 (L^\text{AOD}_\text{recon})_\text{sta} & = \text{dist}\left(\boldsymbol{S}^\text{AOD}_\text{sta}(\boldsymbol{S}^\text{AOD}_\text{sta})^T\boldsymbol{E}_\text{sta}^\text{AOD}, \boldsymbol{E}_\text{sta}^\text{AOD}\right) \\
 L^\text{AOD}_\text{recon} &= \beta(L^\text{AOD}_\text{recon})_\text{dyn} + (1 - \beta)(L^\text{AOD}_\text{recon})_\text{sta}
\label{eq:fun_f}
\end{split}
\end{align}
where $(\boldsymbol{S}^\text{AOD}_\text{dyn})^{T} \in R^{N_\text{dyn}^{\text{S}} \times H \times W} $ and $(\boldsymbol{S}^\text{AOD}_\text{sta})^{T} \in R^{N_\text{sta}^{\text{S}} \times H \times W} $ are the transposes of the AOD-view assignment matrices. $\text{dist}(\cdot,\cdot)$ denotes the $l_2$ norm distance metric. $\beta$ is the importance weight of the dynamic reconstruction loss, and $0\leq\beta\leq1$. Similarly, the meteorology-view reconstruction loss $L^\text{Meteorology}_\text{recon}$ can be obtained. Then, the dual-view reconstruction loss $L_\text{recon}$ is calculated, formulated as:
\begin{align}
L_\text{recon} = \gamma{L^\text{AOD}_\text{recon}} +  (1 - \gamma)L^\text{Meteorology}_\text{recon}
\label{eq:fun_f}
\end{align}
where $\gamma$ is the importance weight of the AOD-view reconstruction loss, and $0\leq\gamma\leq1$.

The air quality estimation task aims to get an air quality image ${\widetilde{\boldsymbol{I}}^t_{\text{Air quality}}}$. For the target grid regions for model evaluation, we aim to decrease the air quality estimation loss $L_{\text {est}}$, formulated as:
\begin{align}
L_{\text {est}}=\frac{1}{|A|} \sum_{i=1}^{|A|} |Y_i-\widetilde{Y}_i|
\label{eq:fun_f}
\end{align}
where $A$ is the set of the target grid regions for model evaluation. $Y$ is the set of ground truth air quality. $\widetilde{Y}$ is the set of estimated air quality.

We jointly optimize the air quality estimation loss with the dual-view reconstruction loss, formulated as:
\begin{align}
L = \lambda{L_{\text {est}}} + (1 - \lambda) L_\text{recon}
\label{eq:fun_f}
\end{align}
where $\lambda$ is the importance weight of the air quality estimation loss, and $0\leq\lambda\leq1$.

\begin{figure}[h]
\centering
\includegraphics[width=1\columnwidth]{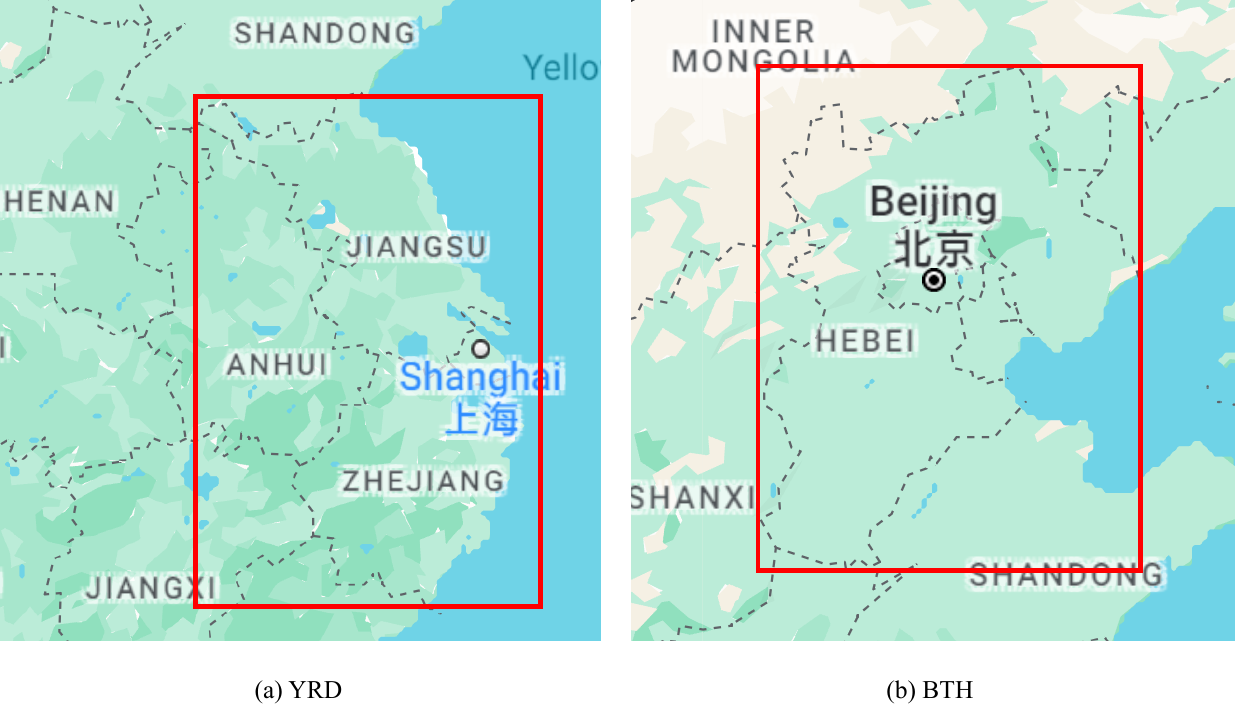} 
\caption{Visualization of the study areas. The study areas are in the red boxes. (Best viewed in color).}
\label{fig9}
\end{figure}

\begin{table*}[t]
\centering
\caption{ Search spaces and the final choices of NNI.}
\label{table1}
\resizebox{\textwidth}{!}{
\begin{tabular}{@{}lccc@{}}
\toprule
\multirow{2}{*}{Hyper-parameter}                   & \multirow{2}{*}{Search space}                   & \multicolumn{2}{c}{Final choices} \\
                                                   &                                                 & YRD-AOD         & BTH-AOD         \\ \midrule
$N^\text{S}_\text{dyn}$: the number of dynamic supergrids                              & $\{2, 3, \dots, 11\}$    & 5              & 6              \\
$N^\text{S}_\text{sta}$: the number of static supergrids                              & $\{2, 3, \dots, 11\}$    & 8              & 10              \\
$\tau$: the historical window length                              & $\{1, 2, \dots, 11\}$    & 6              & 6              \\
$\rho$: threshold $\rho$ in the sparse threshold strategy                              & $\{0.1, 0.2, \dots, 0.9\}$    & 0.4              & 0.4              \\
$\alpha$: the importance weight of the updated AOD-view grid region representation                              & $\{0.1, 0.2, \dots, 0.9\}$    & 0.3              & 0.8              \\
$\beta$: the importance weight of the dynamic reconstruction loss                              & $\{0.1, 0.2, \dots, 0.9\}$    & 0.6              & 0.5              \\
$\gamma$: the importance weight of the AOD-view reconstruction loss                              & $\{0.1, 0.2, \dots, 0.9\}$    & 0.4              & 0.9              \\
$\lambda$: the importance weight of the air quality estimation loss                              & $\{0.1, 0.2, \dots, 0.9\}$    & 0.6              & 0.8              \\
\bottomrule
\end{tabular}}
\end{table*}

\section{Experiments}
In this section, we present extensive experiments to evaluate the performance of DSGNN. We firstly introduce the experimental datasets. Next, we present the experimental settings, the comparison with baselines, the ablation study, the parameter sensitivity analysis, and the case study. 

DSGNN can be used to estimate any kind of air quality data (i.e., $\text{SO}_{2}$, $\text{NO}_{2}$, $\text{PM}_{10}$, $\text{CO}$, $\text{O}_{3}$, or $\text{PM}_{2.5}$). In experiments, the MAE of $\text{PM}_{2.5}$ is used as the performance metric, formulated as:
\begin{equation}
\text { MAE }=\frac{1}{|B|\times|A|} \sum_{i=1}^{|B|}\sum_{j=1}^{|A|}|{P}_{ij}-\widetilde{{P}}_{ij}|
\end{equation}
where $B$ is the set of test samples. $A$ is the set of the target grid regions for model evaluation. ${P}$ is the set of ground truth $\text{PM}_{2.5}$. $\widetilde{{P}}$ is the set of estimated $\text{PM}_{2.5}$.

\subsection{Datasets}
DSGNN is evaluated on two real-world datasets (i.e., YRD-AOD and BTH-AOD), collected from Yangtze River Delta (YRD) and Beijing Tianjin Hebei (BTH), respectively. Both datasets are extracted from 2019/1/1 to 2019/12/31. The study areas of YRD and BTH are show in Figure \ref{fig9}. These datasets include not only the commonly used data (i.e., meteorology data and air quality data) but also AOD data, which have a strong correlation with air pollution concentrations \cite{wei2019satellite}. Recently, Japan Aerospace Exploration Agency released hourly AOD data with 5km spatial resolution during the daytime. However, the AOD data of some grid regions might be missing due to cloud coverage, water glint reflectance, and snow glint reflectance. Following \cite{jiang2021estimation}, we employ modelled AOD data \cite{stafoggia2019estimation} and meteorology data \cite{xiao2017full} as data, and utilize Random Forest \cite{breiman2001random} as the imputation model to impute missing values. The details of these data are as follows:

(1) AOD data are acquired from Japan Aerospace Exploration Agency\footnote{https://www.eorc.jaxa.jp/ptree/index.html}, with the time interval of 1 hour.

(2) Meteorology data (i.e., temperature, humidity, rainfall, wind force, and wind direction) are acquired from the European Centre for Medium-Range Weather Forecasts\footnote{https://cds.climate.copernicus.eu/cdsapp\#!/home}, with the time interval of 1 hour.


(3) Air quality data (i.e., $\text{SO}_{2}$, $\text{NO}_{2}$, $\text{PM}_{10}$, $\text{CO}$, $\text{O}_{3}$, and $\text{PM}_{2.5}$) are acquired from China National Environmental Monitoring Center\footnote{http://www.cnemc.cn/}, with the time interval of 1 hour.

Based on these data, hourly samples are generated. Following \cite{chen2023deep}, we divide YRD and BTH into grid regions, and the size of each grid region is set to 5km × 5km. AOD, meteorology, and air quality images are constructed based on the regional data and grid regions. The width and height of YRD images are 163 and 137, respectively, and the width and height of BTH images are 156 and 151, respectively. The numbers of grid regions with air quality stations are 136 and 65 in YRD images and BTH images, respectively.

\subsection{Experimental Settings}

DSGNN is implemented in Python 3.10.9 with PyTorch 1.13.1, and the source code is released on GitHub\footnote{https://github.com/zx-sunnylife/DSGNN}. The two datasets are split into the training set (70\%), validation set (10\%), and test set (20\%) in chronological order. We aim to estimate the air quality of the target grid regions without air quality stations at a time. Since we only have ground truth for the grid regions with air quality stations, following \cite{chen2023deep}, we randomly divide these grid regions into five equal-sized parts. We train the model five times and average the results as the final performance. Each time we select one different part as the target grid regions for model evaluation, and select the rest parts as model inputs. 

The experiments are conducted on one Nvidia GTX 3080 Ti GPU. Adam is utilized as the optimizer, and the initial learning rate is set to 0.001. The batch size is set to 48. For other hyper-parameters, we exploit the neural network intelligence (NNI) toolkit to automatically select the best choices. The search spaces and the final choices of hyper-parameters are given in Table \ref{table1}. For the configurations of NNI, the max trial number is set to 10 and the optimization algorithm is the Tree-structured Parzen Estimator.

\subsection{Comparison with Baselines}

\begin{table}[]
\centering
\caption{The MAE of different estimation methods. The best results are bolded and the second best results are underlined.}
\label{table2}
\begin{tabular}{lcc}
\toprule
\multicolumn{1}{l}{Method} & \multicolumn{1}{l}{YRD-AOD} & \multicolumn{1}{l}{BTH-AOD} \\ \midrule
LR (CEUS, 2009)                        & 19.42                       & 21.62                       \\
TSRF (ISPRS, 2022)                      & 17.94                       & 19.62                       \\
ANCL (AAAI, 2022)                       & 16.14                       & 19.01                       \\ \midrule
ADAIN (AAAI, 2018)                      & 15.75                       & 18.05                       \\
PANDA (UbiComp, 2019)                      & 15.01                       & 17.11                       \\
MCAM (IJCAI, 2021)                      & 14.13                       & 16.06                       \\
TNSGNN (TSIPN, 2022)                      & 14.69                       & 16.95                       \\ \midrule
Deep-AIR (ACCESS, 2022)                  & 12.47                 & 15.66                        \\
Multi-AP (STTE, 2022)                  & 11.78                 & 14.90                        \\
FAIRY (TCyber, 2023)                     & \underline{11.12}                       & \underline{13.44}                 \\
\textbf{DSGNN (ours)}             & \textbf{8.81}               & \textbf{10.95}              \\      
\bottomrule
\end{tabular}
\end{table}




We compare DSGNN with ten air quality estimation methods, including statistical regression model based methods: LR \cite{shad2009predicting}, TSRF \cite{chen2022urban}, and ANCL \cite{patel2022accurate}, station based deep learning methods: ADAIN \cite{cheng2018neural}, PANDA \cite{chen2019deep}, MCAM \cite{han2021fine}, and TNSGNN \cite{bonet2022explaining}, grid region based deep learning methods: Deep-AIR \cite{zhang2022deep}, Multi-AP \cite{song2022novel}, and FAIRY \cite{chen2023deep}. 

\textbf{Statistical regression model based methods:}

\textbf{LR} \cite{shad2009predicting} uses linear regression for air quality estimation. We train independent models for each province.

\textbf{TSRF} \cite{chen2022urban} employs a nonlinear model to estimate air quality, and the training way is the same as that of LR. This method extracts features from the regional data by Random Forest.

\textbf{ANCL} \cite{patel2022accurate} employs a scalable Gaussian process based model to estimate air quality, and the training way is the same as that of LR. This method utilizes Gaussian processes with feature-specific kernels to extract features from the regional data.

\textbf{Station based deep learning methods:}

\textbf{ADAIN} \cite{cheng2018neural} applies self-attention networks to automatically learn the weights of different air quality stations. This method utilizes FNN to perform automatic feature learning of the regional data.

\textbf{PANDA} \cite{chen2019deep} is a multi-task air quality modeling framework, which employs graph embedding to model the spatial dependencies between the target grid region and other grid regions with air quality stations, and jointly implements air quality estimation and air quality forecasting tasks.

\textbf{MCAM} \cite{han2021fine} employs a multi-channel GNN to model the dynamic and static spatial dependencies between the target grid region and other grid regions with air quality stations.

\textbf{TNSGNN} \cite{bonet2022explaining} introduces a topology-aware kernelized node selection strategy to select the most relevant grid regions with air quality stations for each target grid region, and employs GNN to model the influence of the most relevant grid regions.

\textbf{Grid region based deep learning methods:}

\textbf{Deep-AIR} \cite{zhang2022deep} introduces AirRes module to capture the spatial dependencies of grid regions and provide multi-source feature interactions.

\textbf{Multi-AP} \cite{song2022novel} utilizes FCN \cite{dai2016r} to capture the spatial dependencies of grid regions from micro, meso, and macro views.

\textbf{FAIRY} \cite{chen2023deep} employs SegNet \cite{badrinarayanan2017segnet} to capture the spatial dependencies of grid regions, and uses self-attention networks to model feature interactions.

For statistical regression model based methods and station based deep learning methods, we conduct experiments only based on grid regions with air quality stations. For fairness, all methods follow the same experimental evaluation protocol and use the same regional data (i.e., AOD data and meteorology data). Meanwhile, the parameters of these methods are well tuned.

Table  \ref{table2} presents the overall results of DSGNN and compared baselines on two datasets, from which we can observe the following phenomena:

(1)	In statistical regression model based methods, TSRF and ANCL outperform LR, which demonstrates the effectiveness of modeling the nonlinear relevance between air quality and input data.

(2)	Station based deep learning methods (i.e., ADAIN, PANDA, MCAM, and TNSGNN) outperform TSRF and ANCL. The main reason might be that the performance of statistical regression model based methods depends on feature engineering, which is limited by human knowledge. In contrast, deep learning based methods perform automatic feature learning, which can extract features more comprehensively.

(3)	Grid region based deep learning methods (i.e., Deep-AIR, Multi-AP, and FAIRY) outperform ADAIN, PANDA, MCAM, and TNSGNN, which demonstrates the effectiveness of modeling the spatial dependencies among the target grid regions.

(4)	DSGNN performs better than all baselines, outperforming the best baseline by 20.77\% and 18.52\% in MAE on YRD-AOD and BTH-AOD, respectively. The results justify the advantage of introducing the dual-view supergrid modeling to capture the spatial dependencies of distant grid regions.

    


\subsection{Ablation Study}

To justify the advantage of introducing the dual-view supergrid learning module to generate the AOD-view and meteorolog-view supergrids, we compare DSGNN with several variants. The detailed descriptions of variants are as follows: 

\textbf{DSGNN-CNN (DSGNN-C)} removes the dual-view supergrid learning module and only utilizes a specialized CNN to model the spatial dependencies of adjacent grid regions, which is similar to \textit{G update} and employs the initial grid region representations as input.


\textbf{DSGNN-Dynamic Supergrid (DSGNN-DS)} removes the static supergrids.

\textbf{DSGNN-Static Supergrid (DSGNN-SS)} removes the dynamic supergrids.

\textbf{DSGNN-Low Rank (DSGNN-LR)} removes the sparse threshold strategy and only exploits the low-rank-based mapper in learning assignment matrices.

\textbf{DSGNN-Sparse (DSGNN-S)} removes the low-rank-based mapper and only exploits the sparse threshold strategy in learning assignment matrices.


\textbf{DSGNN-AOD (DSGNN-A)} removes the meteorology-view supergrids.

\textbf{DSGNN-Meteorology (DSGNN-M)} removes the AOD-view supergrids.

Table  \ref{table10} shows the MAE of DSGNN and its variants on YRD-AOD and BTH-AOD, from which we can observe the following phenomena:

(1)	DSGNN outperforms DSGNN-C, which indicates the effectiveness of constructing the supergrid graphs to model the spatial dependencies of distant grid regions.

(2) DSGNN outperforms DSGNN-DS and DSGNN-SS, which justifies the advantage of constructing both the dynamic and static supergrid graphs. By implementing the information interaction on the supergrid graphs (both dynamic and static) and images, DSGNN can model both the dynamic and static spatial dependencies between grid regions.

(3) DSGNN outperforms DSGNN-LR and DSGNN-S, which justifies the advantage of combining the low-rank-based mapper and sparse threshold strategy in learning assignment matrices. In DSGNN, the low-rank-based mapper reduces the computational complexity, while the sparse threshold strategy removes noise and irrelevant information. The incorporation of them enables DSGNN to explore more prominent and informative assignment matrices, leading to superior performance.

(4) DSGNN outperforms DSGNN-A and DSGNN-M, which indicates the advantage of constructing both the AOD-view and meteorology-view supergrid graphs. By implementing the information interaction on the supergrid graphs (both AOD-view and meteorology-view) and images, DSGNN can model the spatial dependencies between grid regions from different views.

To justify the advantage of introducing the implicit correlation encoding module to capture the implicit correlations between pairwise supergrids, we compare DSGNN with several variants. The detailed descriptions of variants are as follows:

\textbf{DSGNN-Sparse Graph (DSGNN-SG)} constructs the supergrid graphs as the sparse graphs based on the top $k$ strategy \cite{wu2020connecting}, and sets top $k$ weights to 1 for each supergrid ($k$ = 3), with the other weights set to 0.

\textbf{DSGNN-Sparse Weighted Graph (DSGNN-SWG)} constructs the supergrid graphs as the sparse graphs based on the top $k$ strategy, and retains top $k$ weights for each supergrid ($k$ = 3), with the other weights set to 0.

\textbf{DSGNN-Fully Connected Graph (DSGNN-FCG)} constructs the supergrid graphs as fully connected graphs, and sets the weight of the implicit correlation to 1.

Table  \ref{table11} shows the MAE of DSGNN and its variants on YRD-AOD and BTH-AOD, from which we can observe the following phenomena:

(1) DSGNN-FCG performs better than DSGNN-SG and DSGNN-SWG, which shows the benefits of constructing the supergrid graphs as fully connected graphs to fully utilize the information between any pairwise supergrids.

(2) DSGNN outperforms DSGNN-FCG, and DSGNN-SWG performs better than DSGNN-SG, which indicates the advantage of introducing the weight of the implicit correlation. By incorporating weights into supergrid graph convolution, DSGNN and DSGNN-SWG can distinguish different implicit correlations.

\begin{table}[]
    \centering
    \caption{the MAE of DSGNN and its variants with different ways to generate supergrids.}
    \label{table10}
    \begin{tabular}{lcc}
    \toprule
    \multicolumn{1}{l}{Method} & \multicolumn{1}{l}{YRD-AOD} & \multicolumn{1}{l}{BTH-AOD} \\ \midrule
    DSGNN-C      & 11.86         & 13.51 \\
    DSGNN-DS                      & 10.09                       & 13.01                       \\
    DSGNN-SS                       & 10.47                       & 13.87                       \\ 
    DSGNN-LR                      & 9.79                       & 12.56                       \\
    DSGNN-S                      & 9.81                       & 12.51                       \\
    DSGNN-A                  & 10.96                 & 12.69                        \\
    DSGNN-M                    & 10.52                       & {12.34}                 \\
    \textbf{DSGNN}             & \textbf{8.81}               & \textbf{10.95}              \\      
    \bottomrule
    \end{tabular}
    \end{table}

    \begin{table}[]
        \centering
        \caption{the MAE of DSGNN and its variants with different ways to capture implicit correlations between pairwise supergrids.}
        \label{table11}
        \begin{tabular}{lcc}
        \toprule
        \multicolumn{1}{l}{Method} & \multicolumn{1}{l}{YRD-AOD} & \multicolumn{1}{l}{BTH-AOD} \\ \midrule
        DSGNN-SG                  & 10.63                 & 12.97                        \\
        DSGNN-SWG                    & {10.26}                       & {12.47}                 \\
        DSGNN-FCG                      & {9.89}                       & {11.90}                 \\
        \textbf{DSGNN}             & \textbf{8.81}               & \textbf{10.95}              \\      
        \bottomrule
        \end{tabular}
        \end{table}

\begin{figure*}[h]
    \centering
    \includegraphics[width=1\textwidth]{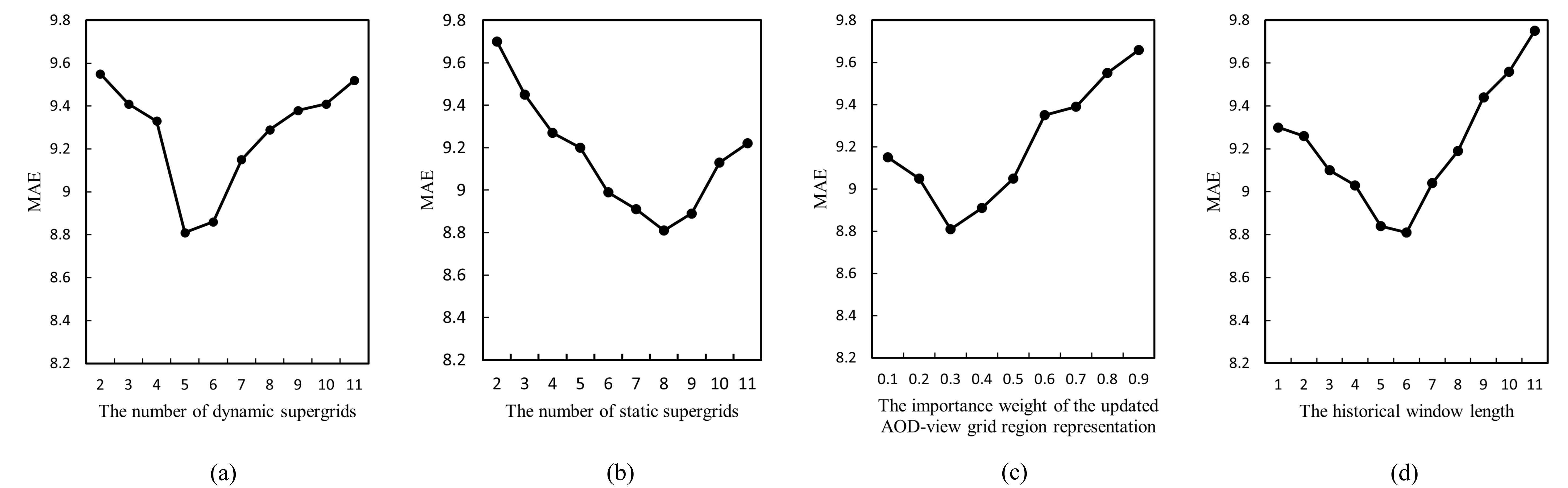} 
    \caption{Impact of hyper-parameters.}
    \label{超参敏感}
\end{figure*}

\begin{figure*}[h]
    \centering
    \includegraphics[width=0.7\textwidth]{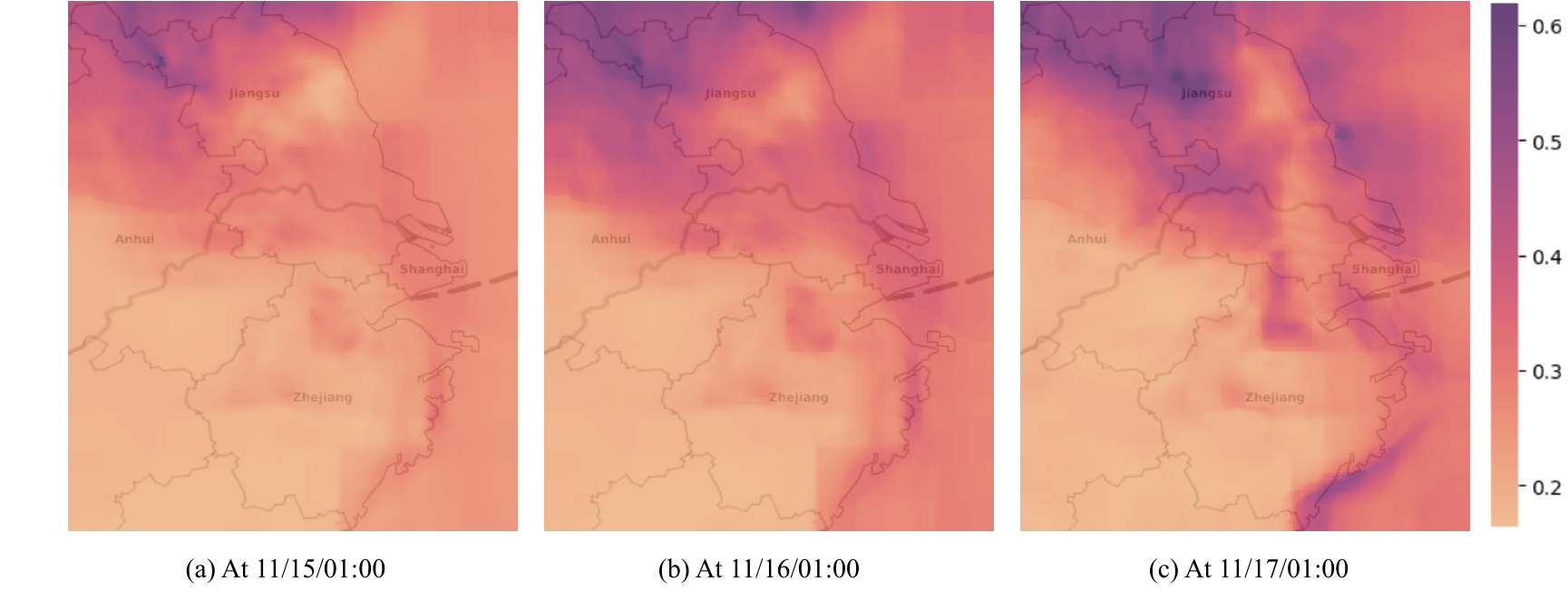} 
    \caption{Visualization of the AOD data at different time steps. (Best viewed in color).}
    \label{case1-不同时刻下的AOD}
\end{figure*}

\begin{figure*}[h]
    \centering
    \includegraphics[width=0.7\textwidth]{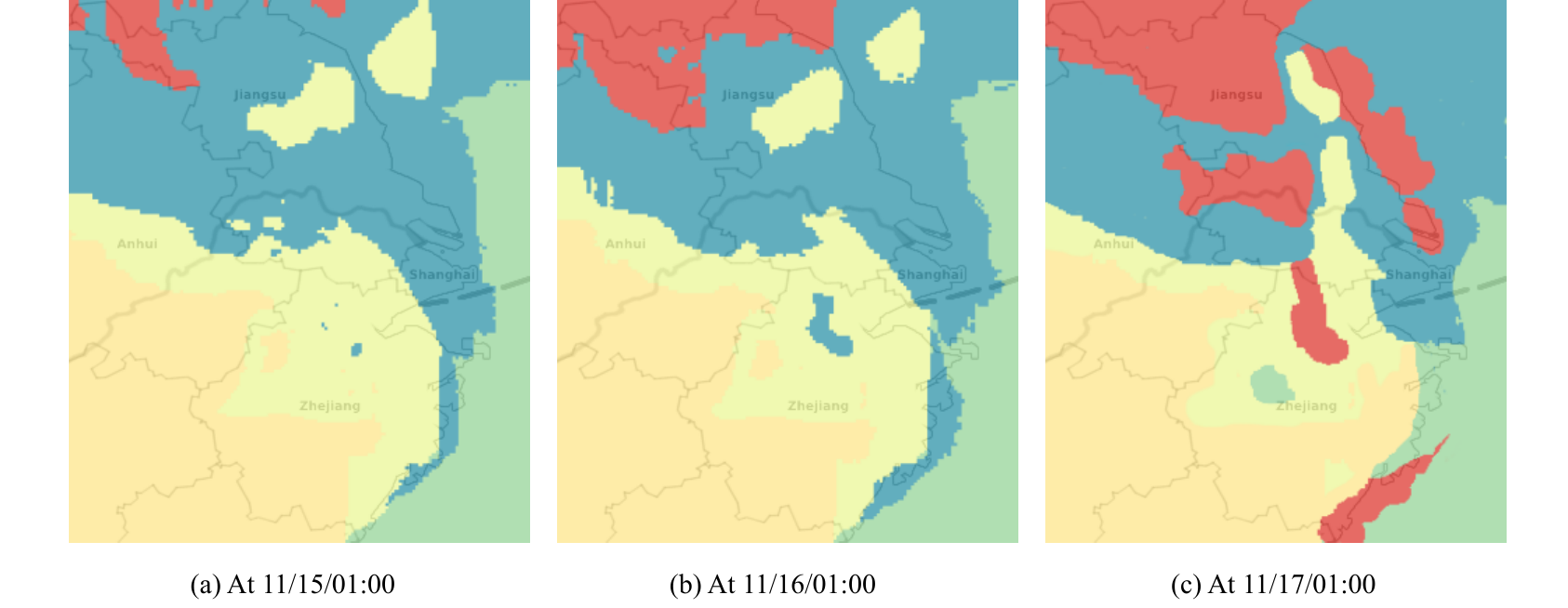} 
    \caption{Visualization of the AOD-view dynamic supergrids at different time steps. (Best viewed in color).}
    \label{case2-不同时刻下的AOD动态超网格}
\end{figure*}

\subsection{Parameter Sensitivity Analysis}
To study the impact of several important parameters, including the number of dynamic supergrids, the number of static supergrids, the importance weight of the updated AOD-view grid region representation, and the historical window length, we conduct the parameter sensitivity analysis on YRD-AOD.

We evaluate the impact of the numbers of supergrids $N^\text{S}_\text{dyn}$ and $N^\text{S}_\text{sta}$, varying them from 2 to 11 with the step of 1. Figure \ref{超参敏感}(a) and 5(b) show the results, from which we can observe that with the increase of the numbers of supergrids, firstly the MAE drops and then rises slightly. The best performances are achieved when the numbers of dynamic and static supergrids are 5 and 8, respectively. The reason may be that the number of edges on the supergrid graph increases exponentially with the number of supergrids, and too many supergrids will bring too many edges and noise, reducing the capability of representing grid regions.



We evaluate the impact of the importance weight of the updated AOD-view grid region representation $\alpha$, varying it from 0.1 to 0.9 with the step of 0.1. Figure \ref{超参敏感}(c) shows the results, from which we can observe that with the increase of the importance weight, firstly the MAE degrades and then increases. The best performance is achieved when the importance weight is 0.3, indicating that the AOD-view spatial dependencies of grid regions are less important than the meteorology-view spatial dependencies.

We evaluate the impact of the historical window length $\tau$, varying it from 1 to 11 with the step of 1. Figure \ref{超参敏感}(d) shows the results, from which we can observe that with the increase of the historical window length, firstly the MAE of DSGNN goes down and then goes up. The best performance is achieved when the historical window length is 6. The reason may be that an excessively long historical window length may bring too many noise and irrelevant information, reducing the capability of representing grid regions.

\subsection{Case Study}



To intuitively reveal the superiority of DSGNN, we conduct several case studies on YRD-AOD.

Figure \ref{case1-不同时刻下的AOD} visualizes the AOD data at three time steps. It can be observed that the AOD data of grid regions are different over time with air diffusing. For example, with the change of time, the dark red grid regions spread from the upper left to the lower right. This indicates that the spatial dependencies between grid regions are different over time. Figure \ref{case2-不同时刻下的AOD动态超网格} visualizes the corresponding AOD-view dynamic supergrids, where different supergrids are distinguished by different colors. Specifically, we mark grid regions by the supergrid that has the highest probability in the assignment matrix. It can be observed that with the change of spatial dependencies between grid regions, the dynamic supergrids contain different grid regions at different time steps. This case illustrates that by grouping correlated grid regions into dynamic supergrids, DSGNN can model the dynamic spatial dependencies between grid regions.

\begin{figure}[h]
    \centering
    \includegraphics[width=1\columnwidth]{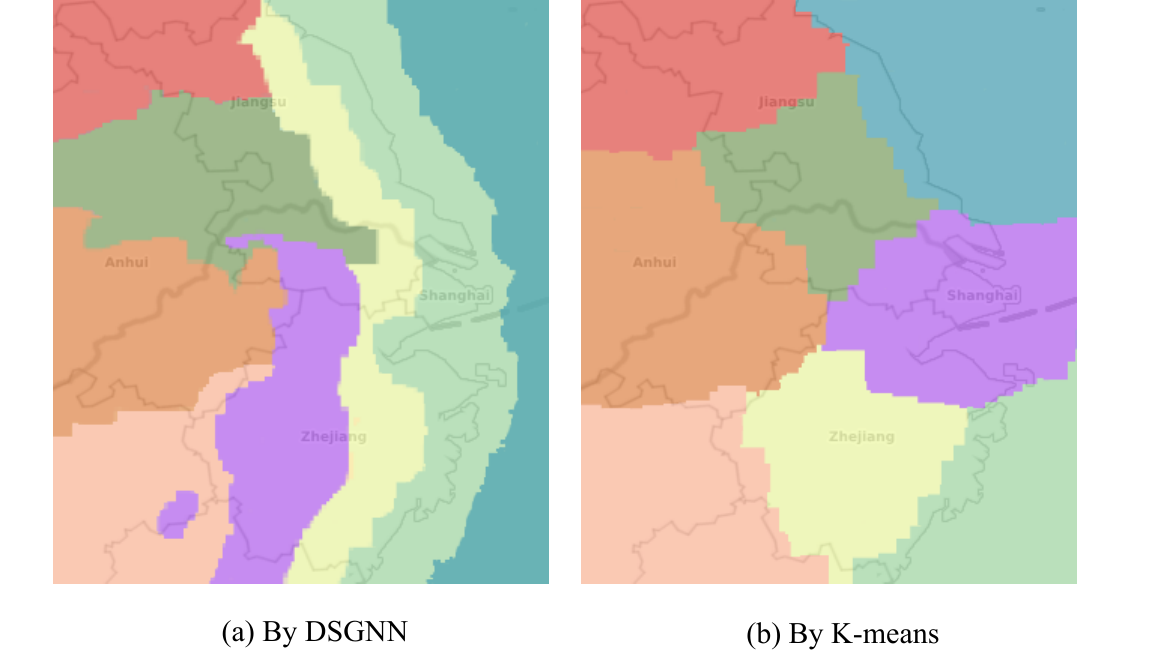} 
    \caption{Visualization of the meteorology-view static supergrids. (Best viewed in color).}
    \label{case3-不同方法得到的气象视角静态超网格}
\end{figure}

Figure \ref{case3-不同方法得到的气象视角静态超网格}(a) visualizes the meteorology-view static supergrids obtained by DSGNN. Figure \ref{case3-不同方法得到的气象视角静态超网格}(b) visualizes the clustering result obtained by K-means based on the meteorology-view static semantic representation ($k = N_\text{sta}^{\text{S}}$). It can be observed that DSGNN can discover some latent static spatial dependencies between grid regions that K-means cannot detect. For example, grid regions distributed along the eastern coast all have a monsoon climate and exhibit strong spatial dependencies, while K-means cannot discover this and groups these grid regions into different supergrids. This case illustrates that by grouping correlated grid regions into static supergrids, DSGNN can better model the static spatial dependencies between grid regions.

\subsection{Conclusions}
Capturing the spatial dependencies of both adjacent and distant grid regions are crucial for the air quality estimation task. To this end, we propose DSGNN, a dual-view supergrid-aware graph neural network for regional air quality estimation. Specifically, we use images to represent the regional data, group grid regions into supergrids from AOD and meteorology views, and construct the dual-view supergrid graphs to model the spatial dependencies of supergrids, which can model the spatial dependencies of both adjacent and distant grid regions. We conduct comprehensive experiments on two real-world datasets and experimental results demonstrate the superiority of DSGNN.

In the future, we will extend this work in the following directions. On the one hand, we plan to extend DSGNN to the air quality forecasting task, which can be realized by designing a forecasting function in the fusion module to obtain the future air quality of all grid regions. On the other hand, we also plan to introduce a multi-level supergrid modeling method to capture more complicated spatial dependencies between grid regions.

\bibliographystyle{IEEEtran}
\bibliography{DSGNN}

\end{document}